%% file: group_rank_judge_acl.tex
\documentclass[11pt]{article}

\usepackage[final]{acl}

\usepackage{times}
\usepackage{latexsym}
\usepackage[T1]{fontenc}
\usepackage[utf8]{inputenc}
\usepackage{microtype}
\usepackage{inconsolata}
\usepackage{graphicx}
\usepackage{booktabs}
\usepackage{amsmath}
\usepackage{amssymb}
\usepackage{amsfonts}
\usepackage{float}
\usepackage{url}
\usepackage{enumitem}
\usepackage[most]{tcolorbox}

\definecolor{prompttitle}{HTML}{17324D}
\definecolor{prompttitlebg}{HTML}{EAF1F8}
\definecolor{promptborder}{HTML}{B7C7D8}
\definecolor{promptfill}{HTML}{FBFCFE}
\definecolor{promptlabel}{HTML}{2F5D7C}
\definecolor{promptinputbg}{HTML}{EAF4FF}
\definecolor{promptinputborder}{HTML}{97BCD9}
\definecolor{promptcodebg}{HTML}{F4F6F8}

\newtcolorbox{promptbox}[1]{
  enhanced,
  breakable,
  colback=promptfill,
  colframe=promptborder,
  colbacktitle=prompttitlebg,
  coltitle=prompttitle,
  boxrule=0.7pt,
  arc=2pt,
  left=8pt,
  right=8pt,
  top=8pt,
  bottom=8pt,
  title={#1},
  fonttitle=\small\bfseries,
  fontupper=\small,
  before skip=6pt,
  after skip=8pt,
}

\newtcolorbox{promptinputblock}{
  enhanced,
  breakable,
  colback=promptinputbg,
  colframe=promptinputborder,
  boxrule=0.5pt,
  arc=1pt,
  left=6pt,
  right=6pt,
  top=5pt,
  bottom=5pt,
  fontupper=\ttfamily\small,
}

\newtcolorbox{promptjsonbox}{
  enhanced,
  breakable,
  colback=promptcodebg,
  colframe=promptborder,
  boxrule=0.5pt,
  arc=1pt,
  left=6pt,
  right=6pt,
  top=5pt,
  bottom=5pt,
  fontupper=\ttfamily\small,
}

\newcommand{\promptrole}[1]{%
  \tcbox[
    on line,
    colback=prompttitlebg,
    colframe=promptborder,
    boxrule=0.5pt,
    arc=1pt,
    left=4pt,
    right=4pt,
    top=1pt,
    bottom=1pt,
  ]{\textbf{#1}}%
}

\newcommand{\promptsection}[1]{%
  \par\medskip\noindent{\color{promptlabel}\textbf{#1}}\par\smallskip%
}

\newcommand{\promptdivider}{%
  \par\medskip{\color{promptborder}\hrule height 0.5pt}\medskip%
}

\title{Prioritizing \textit{the} Best: Incentivizing Reliable Multimodal Reasoning\\by Rewarding Beyond Answer Correctness}

\author{
Mengzhao Jia, Zhihan Zhang, Meng Jiang\\
University of Notre Dame \\
\texttt{mjia2@nd.edu}
}

\begin{document}
\maketitle

\begin{abstract}
\input{parts/0_abstract}
\end{abstract}

\input{parts/1_intro}
\input{parts/2_related}
\input{parts/3_method}
\input{parts/4_exp}

\input{parts/5_conclude}

\bibliography{references}

\appendix
\input{parts/6_appendix}

\end{document}

%% file: parts/0_abstract.tex
Reinforcement Learning with Verifiable Rewards (RLVR) improves multimodal reasoning by rewarding verifiable final answers. Yet answer-correct trajectories may still rely on incomplete derivations, weak evidence, or statements that contradict their conclusions. This gap between answer correctness and reasoning validity, which we call \emph{reasoning-answer inconsistency}, motivates trajectory supervision in multimodal RL. We compare two main approaches: reward models (RMs), and Generative Rewards (GRs). RMs are efficient and help early in training, but their gains weaken as the policy distribution shifts; GRs improve performance, but may give unstable rewards and computationally expensive. We therefore propose \textbf{Groupwise Ranking Reward}, which ranks verifier-passed trajectories for the same prompt in one pass and redistributes reward accordingly. Groupwise comparison better separates stronger and weaker correct trajectories with lower judge overhead than GRs. Experiments show that RLVR aggravates reasoning-answer inconsistency, while trajectory supervision alleviates it. Groupwise Ranking Reward performs best overall, improving reliability-conditioned accuracy from 47.4\% to 54.7\% over RLVR.

%% file: parts/1_intro.tex
\section{Introduction}
\label{sec:intro}

\begin{figure*}[t]
\centering
\includegraphics[width=\textwidth,keepaspectratio]{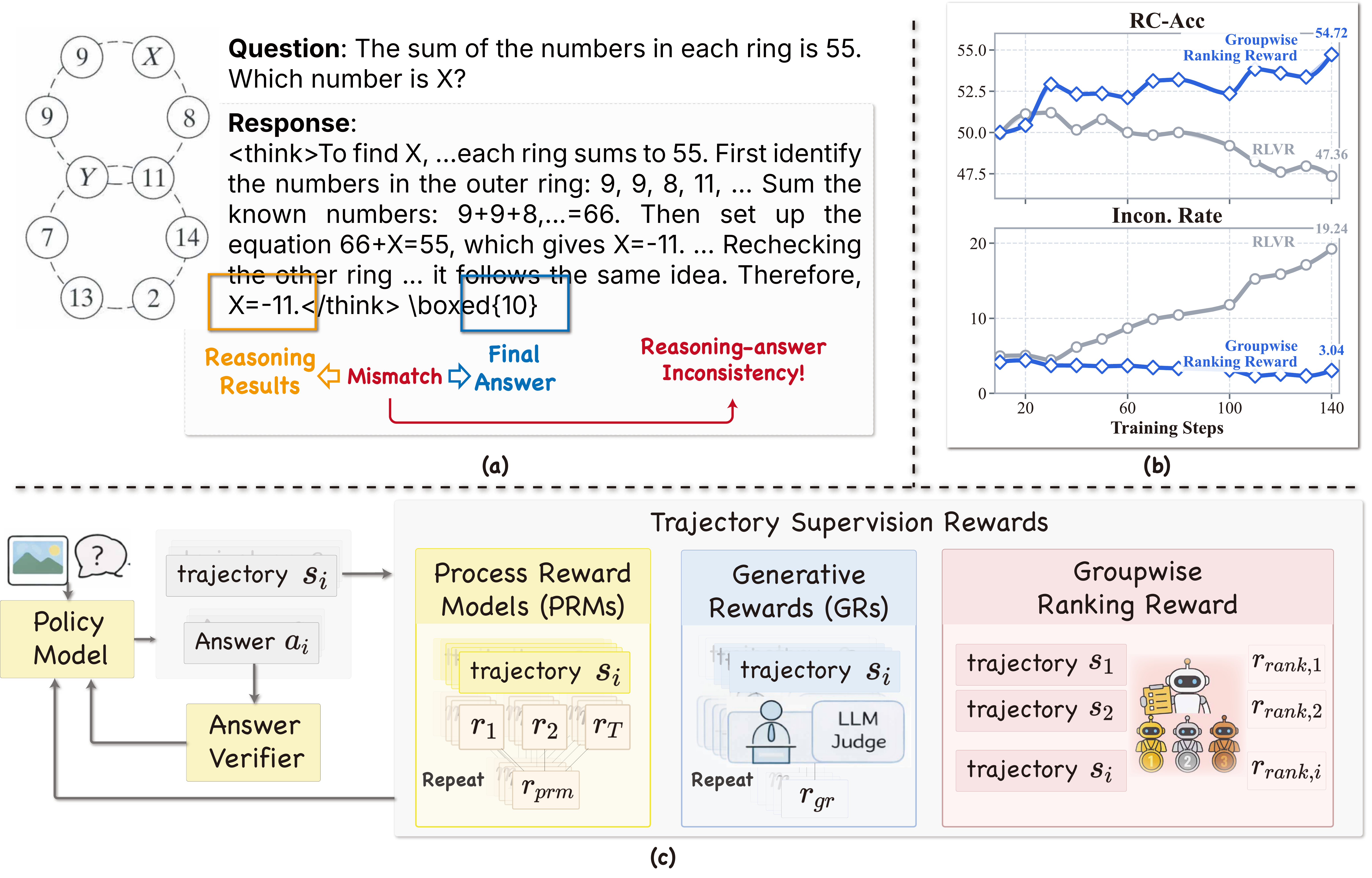}
\caption{\textbf{Reasoning-answer inconsistency under outcome-only RLVR and the motivation for Groupwise Ranking Reward.} (a) An example of reasoning-answer inconsistency: the trajectory derives \(X=-11\) in its reasoning, but the final boxed answer is \(10\). (b) Training dynamics under outcome-only RLVR and Groupwise Ranking Reward. Outcome-only RLVR steadily increases inconsistency and eventually reduces reliability-conditioned accuracy (\(\mathrm{RC\text{-}Acc}\)), whereas Groupwise Ranking Reward keeps inconsistency stable and continues improving \(\mathrm{RC\text{-}Acc}\). (c) Comparison of trajectory supervision schemes. PRMs and pointwise GRs score each trajectory independently, while Groupwise Ranking Reward compares verifier-passed trajectories for the same prompt jointly and assigns ranking-based rewards according to their relative quality.}
\label{fig:main_placeholder}
\end{figure*}

Multimodal Large Language Models (MLLMs) are increasingly capable of visually grounded reasoning  \citep{shi2024mathllava,wang2025mathcodervl}, and Reinforcement Learning with Verifiable Rewards (RLVR)~\cite{guo2025deepseekr1} has become a standard way to push this progress further by directly optimizing verifiable outcomes such as final answer correctness  \citep{zhang2025openmmreasoner,wang2025internvl35,wei2025openvisionreasoner}. 
At the same time, recent studies have noted that under outcome-only supervision, MLLMs' reasoning process can become less reliable as training continues, even when the final answer is correct \citep{chen2025grpocare,kan2025taco,huang2025acre,wang2025scs,arxiv2510_14738}. A closer look at the training process shows that answer-correct responses increasingly rely on incomplete derivations, weak evidence, or even statements that conflict with their own conclusions. As shown in Figure~\ref{fig:main_placeholder}(a): the reasoning trajectory derives \(X=-11\), but the final boxed answer is \(10\), the response would still be rewarded by outcome-only supervision despite an unfaithful reasoning chain. We trace this problem to a basic limitation of outcome-only RLVR: it supervises whether the final answer is correct, but not whether the intermediate reasoning process actually justifies that answer. We refer to this failure mode as \textit{reasoning-answer inconsistency}.  Left unresolved, this failure mode can make MLLMs unreliable in applications such as medical decision support, autonomous driving, and scientific analysis, where a correct-looking answer with contradictory reasoning is not acceptable.

A natural way to address this problem is to add supervision over the validity of the reasoning trajectory. However, how to introduce such supervision signals into multimodal RL in a way that is both effective and computationally efficient remains underexplored. In this paper, we compare two main families of trajectory supervision methods: \textbf{Reward Models} (RMs) and \textbf{Generative Rewards} (GRs).
RMs directly assign scalar quality scores to the reasoning process. We use the fine-grained \textbf{Process Reward Model} (PRM) that assigns a score to each intermediate reasoning step  \citep{uesato2022solving,lightman2023lets,arxiv2312_08935}. 
On the other hand, GRs use a Large Language Model (LLM) to follow a predefined evaluation instruction, generate a textual judgment then assign a scalar score   \citep{zheng2023judging,arxiv2401_10020,arxiv2503_03064}. 
Our experiments show that RMs are computationally efficient and help in the early stages of training, but their benefits weaken as the policy distribution shifts; GRs improve performance to some extent, but their gains are limited by instability and come with severe efficiency costs.
Based on these observations, we propose a \textbf{Groupwise Ranking Reward}, which ranks verifier-passed trajectories of the same problem in a single pass, and redistributes rewards based on that ranking. Figure~\ref{fig:main_placeholder}(c) highlights the key distinction: PRMs and pointwise GRs score each trajectory independently, whereas our method compares multiple trajectories jointly and rewards them according to relative quality. By using groupwise comparison instead of independent scoring, the method better distinguishes stronger from weaker correct trajectories with lower judge computation overhead than GRs. It achieves the strongest performance among all trajectory supervision methods tested.

We conduct extensive experiments with different judge models, reward designs, and training data to systematically analyze this problem and evaluate the proposed method. The results show that RLVR can aggravate reasoning-answer inconsistency, whereas trajectory supervision alleviates this effect. Figure~\ref{fig:main_placeholder}(b) previews this trend: under outcome-only RLVR, inconsistency rises over training and eventually reduces Reliability-Conditioned Accuracy ($\mathrm{RC\text{-}Acc}$), while Groupwise Ranking Reward keeps inconsistency stable and continues improving \(\mathrm{RC\text{-}Acc}\). Among the approaches we study, the Groupwise Ranking Reward achieves both the best accuracy and reasoning faithfulness, improving $\mathrm{RC\text{-}Acc}$ from 47.4\% to 54.7\% over RLVR. These gains indicate that even when answer correctness is already verified, multimodal RL still lacks a way to prefer better-grounded correct trajectories.

%% file: parts/2_related.tex
\section{Related Work}

\textbf{Reinforcement Learning for Multimodal Reasoning}. Chain-of-Thought prompting and RLHF laid the foundation for training LLMs to produce explicit multi-step reasoning \citep{arxiv2201_11903,arxiv2203_02155}. Building on this trend, recent MLLMs extend stepwise reasoning to visually grounded tasks through reasoning-oriented supervision and post-training, especially in multimodal mathematical reasoning and general-purpose visual reasoning \citep{shi2024mathllava,wang2025mathcodervl,zhang2025openmmreasoner,wang2025internvl35,wei2025openvisionreasoner}. RLVR has become a natural recipe in this setting because verifiable rewards allow direct optimization of answer correctness without requiring dense trajectory annotations, and recent pipelines such as GRPO and DeepSeek-R1 show that this paradigm can scale beyond narrow domains \citep{shao2024deepseekmath,guo2025deepseekr1,arxiv2503_23829}. Our work complements these efforts by studying how RLVR should assign differential credit among verifier-passed rollouts, encouraging the policy to prefer better-grounded and more reliable reasoning trajectories.

\textbf{Reasoning-Answer Inconsistency in MLLMs}. Recent multimodal RL work identifies reasoning-answer inconsistency as a common side effect of outcome-only RLVR, where answer accuracy can improve even as answer-reasoning coherence degrades. Existing fixes mainly follow three directions. One uses reference-policy calibration \citep{chen2025grpocare,kan2025taco}, but this signal becomes less reliable as the online policy drifts. Another enforces consistency under semantics-preserving perturbations \citep{huang2025acre,wang2025scs}, though these methods are most natural for closed-answer settings. A third line uses rubric-based generative rewards \citep{arxiv2510_14738}, which provide richer supervision but introduce substantial judge overhead without being updated jointly with the evolving policy.

\textbf{Trajectory Supervision}. A parallel line of work improves reasoning by supervising intermediate steps rather than only final answers. Trajectory supervision was first studied in text-only mathematical reasoning \citep{uesato2022solving,lightman2023lets} and has recently been extended to multimodal settings \citep{arxiv2508_04088}. Another line of work uses language models themselves as evaluators or reward models \citep{zheng2023judging,arxiv2401_10020,arxiv2411_15594}. Recent work further shows that listwise or distribution-aware judge inference can be more reliable than independently assigning a single absolute score to each candidate \citep{arxiv2410_04346,arxiv2503_03064}. Closely related, RLRR replaces absolute reward shaping with relative ranking in group-based RL and trains a ranking reward model to produce groupwise preferences \citep{arxiv2601_23058}. However, that work mainly uses ranking to improve reward modeling and optimization. Our method aligns with these directions: we study the effects of both RMs and GRs for trajectory supervision in multimodal RL. Moreover, we propose groupwise ranking methods aimed at improving the reliability of multimodal reasoning.

%% file: parts/3_method.tex
\section{Method}

\label{sec:method}

RLVR improves answer correctness, but it does not distinguish among verifier-passed trajectories whose reasoning quality differs substantially. As a result, the policy model can still be rewarded for correct answers supported by incomplete or contradictory traces. 
To address this limitation, we study three trajectory supervision reward designs within the same RLVR pipeline. The first is a \textbf{Reward Model} (RM), which assigns scalar scores; in our experiments, this RM is instantiated as a fine-grained \textbf{Process Reward Model} (PRM) that scores intermediate reasoning steps \citep{uesato2022solving,lightman2023lets,arxiv2312_08935,arxiv2508_04088}. The second is a standard \textbf{Generative Reward} (GR), which uses an LLM-as-a-Judge to produce a textual judgment and score each trajectory independently \citep{zheng2023judging,arxiv2401_10020,arxiv2411_15594}. The third is our \textbf{Groupwise Ranking Reward}, a groupwise variant of generative reward that compares verifier-passed trajectories for the same prompt and redistributes reward according to their relative quality ranking \citep{arxiv2410_04346,arxiv2503_03064}.

\subsection{Problem Setup}
\label{subsec:prelim}

Let $\mathcal{D}$ be a multimodal training set. Each training example is denoted by $(x, a^*)$, where $x=(v,q)$, $v$ is the input image, $q$ is the textual question, and $a^*$ is the ground-truth answer. Given $x$, the policy $\pi_\theta(\cdot \mid x)$ generates a structured response
\begin{equation}
    y = [z, a],
\end{equation}
where $z$ is the reasoning trajectory enclosed in \texttt{<think>...</think>} and $a$ is the final answer extracted from the \texttt{\textbackslash boxed\{...\}} expression.
In the RLVR pipeline, each sampled complete response $y=[z,a]$ is referred to as a \emph{rollout}.

Following standard RLVR pipelines for reasoning \citep{shao2024deepseekmath,guo2025deepseekr1}, a rule-based verifier provides a deterministic verification reward
\begin{equation}
    r_{\text{ver}}(y) = \mathbb{I}[\operatorname{Verify}(a, a^*) = 1],
\end{equation}
where $\mathbb{I}[\cdot]$ is the indicator function. RLVR maximizes $\mathbb{E}_{(x, a^*) \sim \mathcal{D}, y \sim \pi_\theta}[r_{\text{ver}}(y)]$. This objective gives identical reward to all verifier-passed responses; however, their reasoning quality may differ substantially.
We use the standard GRPO \citep{shao2024deepseekmath} optimization pipeline throughout: for each prompt $x$, we sample a rollout group $Y=\{y_i=[z_i,a_i]\}_{i=1}^N$ from $\pi_\theta(\cdot \mid x)$ and optimize over this group using verifier and trajectory rewards. Here we focus on how different methods define the trajectory supervision signal; the GRPO optimization objective and equations are given in Appendix~\ref{app:standard_grpo}.

\subsection{Trajectory Reward Variants}
\label{subsec:judge}

Beyond the verifier reward $r_{\text{ver}}(y_i)$, we add an auxiliary trajectory reward $r_{\text{aux}, i}$ to provide extra supervision on rollout $y_i=[z_i, a_i]$. Common trajectory rewards include RMs and GRs. 

\paragraph{RMs.}
RMs assign scalar quality scores without generating extra natural language justifications. They can operate at different supervision granularities. In this paper, we instantiate the RM as a fine-grained PRM that decomposes the reasoning trajectory into intermediate steps and assigns a step-level quality signal along the trajectory. Let $\mathcal{S}_{\text{prm}}$ denote the reward model and let $\{s^{\text{prm}}_{i,t}\}_{t=1}^{T_i} = \mathcal{S}_{\text{prm}}(x, z_i)$ be the scores for the $T_i$ reasoning steps in rollout $i$. The rollout-level PRM reward is obtained by aggregating these step scores,
\begin{equation}
    r_{\text{prm}, i} = \operatorname{Agg}\!\left(\{s^{\text{prm}}_{i,t}\}_{t=1}^{T_i}\right),
\end{equation}
where $\operatorname{Agg}(\cdot)$ denotes the aggregation rule. PRM provides dense supervision over the reasoning process.

\paragraph{GRs.}
GRs use an LLM judge that follows an evaluation instruction with explicit scoring rubrics. For each rollout, the judge generates a textual judgment together with a rubric-grounded scalar score, from which the reward is parsed:
\begin{equation}
    r_{\text{gr}, i} = \operatorname{ParseScore}\!\left(\mathcal{S}_{\text{gr}}(q, z_i, a_i)\right).
\end{equation}
Compared with RMs, GRs do not directly emit scores; instead, the judge first produces an evaluative judgment, and the scalar reward is then extracted from that judgment. 
In our setup, the rubric casts the judge as an expert evaluator of reasoning quality, defines anchored score levels in $[0,1]$, and asks it to assess dimensions such as logical coherence, calculation quality, completeness of the conclusion, and honest handling of uncertainty. The judge returns a short feedback string together with a JSON-formatted scalar score, which we parse as the reward. We refer to the setting where each rollout is judged independently as \emph{pointwise GR}, to distinguish it from the \emph{groupwise GR} setting introduced next.

RMs and GRs are both widely used trajectory supervision signals, but they also have limitations: PRM can weaken under policy shift, while pointwise GR depends on the absolute score calibration and requires separate judging for each rollout. Motivated by this, we further introduce Groupwise Ranking Reward as a groupwise variant of GRs.

\paragraph{Groupwise Ranking Reward.}
Groupwise Ranking Reward extends GRs from independent scoring to within-group ranking. For each prompt $x$, we keep only the verifier-passed subset $I^+=\{i:r_{\text{ver}}(y_i)=1\}$ within the sampled rollout group $Y$, where $|I^+|$ denotes the number of verifier-passed rollouts, because incorrect rollouts are already handled by the verifier. We focus on finding which correct rollout is best in reliability and quality.

The groupwise judge $\mathcal{S}_{\text{cmp}}$ takes the shared question $q$ together with all verifier-passed reasoning-answer pairs $\{(z_i, a_i)\}_{i \in I^+}$, compares them side-by-side, and outputs $M$ ordered tie-aware tiers
\begin{equation}
    (\tau_1, \tau_2, \dots, \tau_M) = \mathcal{S}_{\text{cmp}}\!\left(q, \{(z_i, a_i)\}_{i \in I^+}\right),
\end{equation}
where $\tau_1 \succ \tau_2 \succ \cdots \succ \tau_M$, $\tau_m$ contains the candidates assigned to the $m$-th rank tier, and $\tau_1$ is the top tier. 

We design a rank-to-score conversion based on a simple intuition: a rollout should receive a higher score if it is ranked ahead of more verifier-passed rollouts, and a lower score if it is ranked ahead of fewer. We also preserve ties instead of forcing arbitrary tie-breaking, avoiding injecting artificial noise when the judge views several traces as equally rigorous. Concretely, for any rollout $i \in \tau_m$, let $b_i=\sum_{\ell=m+1}^{M} |\tau_\ell|$ be the number of lower-ranked correct rollouts and let $c_i=|\tau_m|-1$ be the number of tied rollouts. The raw score is
\begin{equation}
    \tilde{s}_i = \frac{b_i + \frac{1}{2} c_i}{|I^+|-1},
\end{equation}
where each lower-ranked rollout contributes one point, each tie contributes half a point, and the result is normalized by the number of available comparisons. 
Thus, candidates in $\tau_1$ receive the highest raw score and candidates in $\tau_M$ the lowest, while prompts with different $|I^+|$ remain on the same $[0,1]$ scale. This normalization makes the rank-to-score mapping well-defined for prompt groups with any number of verifier-passed rollouts.

The final ranking reward is obtained by centering these raw scores within the verifier-passed set. Let $\bar{s}=\frac{1}{|I^+|}\sum_{j \in I^+} \tilde{s}_j$ denote the mean raw score over $I^+$.
\begin{equation}
    r_{\text{rank}, i} =
    \begin{cases}
      \tilde{s}_i - \bar{s}, & \text{if } i \in I^+,\, |I^+| \ge 2, \\
      0, & \text{otherwise}.
    \end{cases}
\end{equation}
This centering makes the ranking reward zero-mean within the verifier-passed set. Under this mapping, $r_{\text{rank}, i}$ is bounded in $[-0.5, 0.5]$. Thus, the judge acts as a pure redistribution mechanism: better-supported correct rollouts receive positive ranking reward, while weaker correct rollouts can still have positive verifier reward but negative ranking reward, which effectively penalizes weaker correct reasoning among answer-correct rollouts. Incorrect rollouts remain untouched and still receive zero total reward. If all verifier-passed rollouts fall into the same tier, the centered reward is again zero. Appendix~\ref{Groupwise_Ranking_Reward} gives the full implementation details and summarizes alternative rank-to-score mappings.

\subsection{Training Objective}
\label{subsec:optimization}

The final rollout reward combines verification with one auxiliary trajectory reward:
\begin{equation}
    \mathcal{R}_i = r_{\text{ver}}(y_i) + \lambda r_{\text{aux}, i},
\end{equation}
where $\lambda \ge 0$ controls the strength of trajectory supervision and $r_{\text{aux}, i} \in \{r_{\text{prm}, i}, r_{\text{gr}, i}, r_{\text{rank}, i}\}$. In our experiments, we use $\lambda=1$ for all three variants. The optimization algorithm is unchanged.

%% file: parts/4_exp.tex
\section{Experiments}
\label{sec:exp}

\newcommand{\resultcell}[3]{\shortstack[c]{#1 \\ {\footnotesize #2 / #3}}}
\newcommand{\methodcell}[1]{\shortstack[l]{#1}}

\input{parts/4_exp/4_1_setup}
\input{parts/4_exp/4_2_main_results}
\input{parts/4_exp/4_3_analysis}

%% file: parts/4_exp/4_1_setup.tex
\subsection{Experimental Setup}

\begin{table*}[t]
\centering
\caption{Model performance comparison results under a matched training budget. The second and third columns report five-benchmark average \(\mathrm{RC\text{-}Acc}\) and \(\mathrm{Acc}\), respectively, while the remaining benchmark columns report \(\mathrm{RC\text{-}Acc}\). Groupwise Ranking Reward achieves the best overall performance, yielding the highest average \(\mathrm{RC\text{-}Acc}\) and leading on four of the five benchmarks. We bold the best value in each numeric column and underline the second best.}
\small
\setlength{\tabcolsep}{3.6pt}
\renewcommand{\arraystretch}{1.15}
\begin{tabular}{lccccccc}
\toprule
\textbf{Method}
& \textbf{RC-Acc Avg.}
& \textbf{Acc Avg.}
& \textbf{MathVision}
& \textbf{MathVista}
& \textbf{MMMU}
& \textbf{MMMU-Pro}
& \textbf{WeMath} \\
\midrule
\multicolumn{8}{c}{\emph{Reference checkpoints}} \\
\midrule
\methodcell{Qwen2.5-VL-7B-IT}
& 47.3
& 49.0
& 24.8
& 67.2
& 52.6
& 35.2
& 56.6 \\

\methodcell{RLVR}
& 47.4
& 53.6
& 22.8
& 70.8
& 48.2
& 36.2
& 58.8 \\

\midrule
\multicolumn{8}{c}{\emph{Controlled comparison with different reward methods}} \\
\midrule
\methodcell{\textit{w/} PRM}
& 46.8
& 49.1
& 26.4
& 66.4
& 50.0
& 30.6
& 60.8 \\

\methodcell{\textit{w/} Pointwise GR}
& \underline{52.3}
& \underline{53.8}
& \underline{31.0}
& \underline{71.0}
& \underline{56.6}
& \textbf{39.0}
& \underline{63.8} \\

\methodcell{\textit{w/} Groupwise Ranking Reward}
& \textbf{54.7}
& \textbf{55.9}
& \textbf{31.4}
& \textbf{74.8}
& \textbf{57.8}
& \underline{38.8}
& \textbf{70.8} \\
\bottomrule
\end{tabular}
\vspace{-0.35em}
\label{tab:main_step140}
\end{table*}

\noindent\textbf{Training Hyperparameters.} All RL training runs start from the save base checkpoint, \nolinkurl{Qwen2.5-VL-7B-Instruct} \citep{bai2025qwen25vl}. Across all RL variants, we use rollout budget \(n=8\), auxiliary-reward coefficient \(\lambda=1\), learning rate \(1\times 10^{-6}\), KL coefficient \(1\times 10^{-2}\). All experiments use the same \textit{ViRL} \citep{wang2025vlrethinker} training data and are run for 2 epochs.

\noindent\textbf{Methods to Compare.} Our main experiment compares four RL variants on \textit{ViRL}: outcome-only RLVR, \textit{w/} PRM, \textit{w/} Pointwise GR, and \textit{w/} Groupwise Ranking Reward. All four use the same GRPO optimization pipeline and differ only in how trajectory supervision is added. The PRM variant uses a visual process-reward model, Qwen-VL-PRM-7B~\cite{qwen-prm}, that scores each trajectory step with a binary true/false decision and averages rewards across all steps. The two GR-based variants both use \texttt{gpt-oss-20b} \citep{openai2025gptosscard} as the judge over textualized question-trajectory-answer tuples: \textit{w/} Pointwise GR gives a scalar \texttt{judge\_score} in \([0,1]\), while \textit{w/} Groupwise Ranking Reward assigns tie-aware ranks for verifier-passed candidates and maps them to centered groupwise rewards. Appendix~\ref{app:reward_judge_instructions} lists the judge prompts.

\noindent\textbf{Benchmarks.} We evaluate final checkpoints on five multimodal reasoning benchmarks: three math-focused benchmarks, MathVision \citep{wang2024mathvision}, MathVista \citep{lu2024mathvista}, and WeMath \citep{qiao2025wemath}, and two broader general-reasoning benchmarks, MMMU \citep{yue2024mmmu} and MMMU-Pro \citep{yue2024mmmupro}. 
To make the reported average reasoning–answer inconsistency across all benchmarks fair, we randomly sample 500 instances from each dataset to construct our final test set. In this way, each benchmark contributes equally to the overall average; otherwise, we find the value could be artificially inflated.

\noindent\textbf{Evaluation Metrics.}\label{sec:stricacc} The standard answer Accuracy (\(\mathrm{Acc}\)) checks only whether the final answer matches the ground truth. In our setting, this is not sufficient because we frequently observe \emph{reasoning-answer inconsistency}: the final answer is correct, but the trajectory supports a different conclusion or does not justify the boxed answer. Such predictions are hard to trust even when they count as correct under the standard Accuracy metric. To measure this failure mode, we ask a \emph{judge model} to read the whole trajectory, including the final answer, and generate a binary consistency score. We then report the overall inconsistency rate (\(\mathrm{IncR}\)), the correct-but-inconsistent rate (\(\mathrm{CBIR}\), the final answer is correct but the trajectory fails to pass the inconsistency judgment), and Reliability-Conditioned Accuracy (\(\mathrm{RC\text{-}Acc}\)) which is defined as
$\mathrm{RC\text{-}Acc}=\mathrm{Acc}-\mathrm{CBIR}$.
That is to say, \(\mathrm{RC\text{-}Acc}\) removes answer-correct but reasoning-inconsistent cases, making a stricter criteria, thus we use it as the primary metric throughout the paper.

\noindent\textbf{Reasoning-Answer Inconsistency Judge.} When judging whether a trajectory is inconsistent, we decompose the task into four simple steps: (i) extract the conclusion implied by the trajectory, (ii) extract and normalize the final boxed answer, including mapping option letters to option content when the prompt provides it, (iii) compare the two under lightweight equivalence rules such as unit omission and numeric reformatting, and (iv) abstain with N/A when the model did not properly provide a valid answer (abstentions do not count toward inconsistency rate). This decomposition makes the intermediate decision points explicit and easier to audit.

%% file: parts/4_exp/4_2_main_results.tex
\subsection{Main Results}

Table~\ref{tab:main_step140} reports the five-benchmark average \(\mathrm{RC\text{-}Acc}\) and \(\mathrm{Acc}\), together with the benchmark-wise \(\mathrm{RC\text{-}Acc}\), after training with different methods for 2 epochs. Appendix Table~\ref{tab:main_step140_detailed} provides the complete scores, including \(\mathrm{Acc}\), \(\mathrm{RC\text{-}Acc}\), and \(\mathrm{CBIR}\).

\textbf{RLVR improves answer accuracy but not reasoning reliability.} Compared to the base checkpoint, RLVR raises average answer accuracy, but its  \(\mathrm{RC\text{-}Acc}\) remains unchanged because correct-but-inconsistent predictions become much more frequent (see Table ~\ref{tab:main_step140_detailed} for \(\mathrm{CBIR}\) scores). This reveals a significant side effect of outcome-only RL: the policy gets the final answer right more often, but the reward still cannot tell whether the trajectory actually supports that answer. \textbf{GR-based rewards recover this gap, and Groupwise Ranking Reward gives the best overall trade-off.} Under the same training budget, both GR-based variants outperform RLVR and the PRM variant, and Groupwise Ranking Reward reaches the strongest average \(\mathrm{RC\text{-}Acc}\)  at \(54.7\%\), compared with \(52.3\%\) for Pointwise GR and \(47.4\%\) for RLVR. This suggests that comparing correct candidates within the same prompt gives a more reliable signal than scoring each rollout in isolation. \textbf{The gains are broad rather than benchmark-specific.} Groupwise Ranking Reward performs best on four of the five benchmarks and remains competitive on MMMU-Pro, with especially large gains on WeMath. The improvement is therefore not tied to one dataset or one answer format.

\begin{figure*}[t]
\centering
\includegraphics[width=\textwidth]{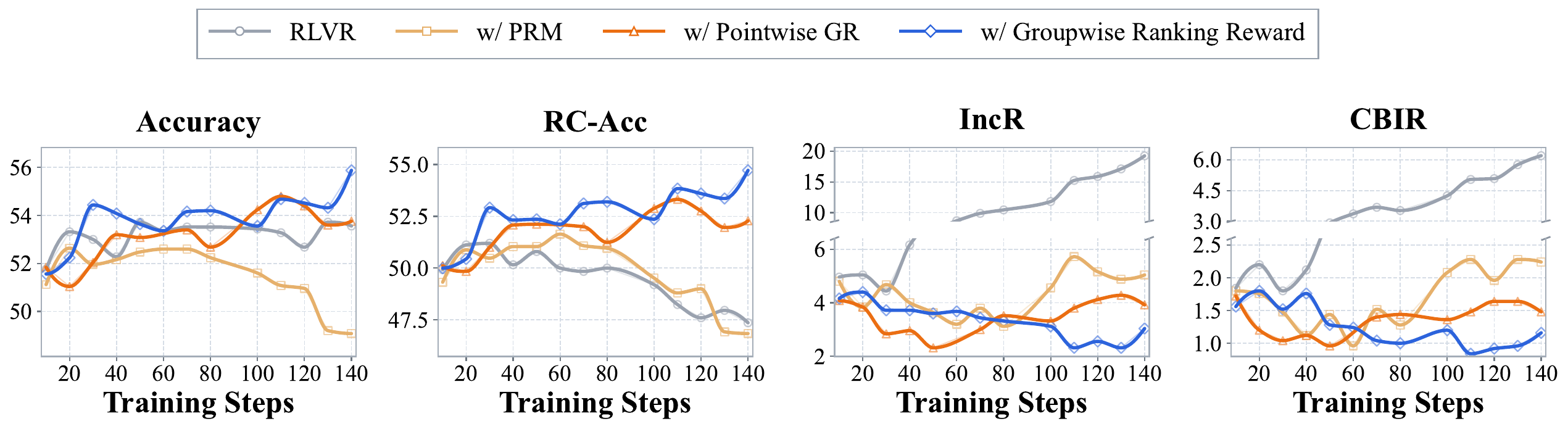}
\caption{Training dynamics across reward variants. The four panels report Accuracy, RC-Acc, IncR, and CBIR, and all reported values are percentages. The last two panels use a broken y-axis so the large RLVR values remain visible while differences among the lower-valued methods are still readable. Although outcome-only RL can improve final-answer correctness, it simultaneously degrades reasoning reliability in multimodal reasoning tasks. By contrast, Groupwise Ranking Reward is the only method that consistently improves both answer accuracy and \(\mathrm{RC\text{-}Acc}\) while simultaneously reducing \(\mathrm{IncR}\) and \(\mathrm{CBIR}\).}
\label{fig:main_step_curves}
\end{figure*}

%% file: parts/4_exp/4_3_analysis.tex
\subsection{Analysis}

\input{parts/4_exp/4_3a_training_dynamics}

\input{parts/4_exp/4_3b_model_size}
\input{parts/4_exp/4_3c_mapping_function}
\input{parts/4_exp/4_3d_stability}

%% file: parts/4_exp/4_3a_training_dynamics.tex
\subsubsection{Training Dynamics}
Figure~\ref{fig:main_step_curves} shows the five-benchmark average trends for \(\mathrm{Acc}\), \(\mathrm{RC\text{-}Acc}\), \(\mathrm{IncR}\), and \(\mathrm{CBIR}\), so we can see how accuracy and inconsistency evolve throughout training rather than only at the final checkpoint.


\textbf{Outcome-only reward harms reasoning reliability as training progresses.}
Although RLVR improves standard answer accuracy from \(51\%\) to \(54\%\), as shown in the first panel of Figure~\ref{fig:main_step_curves}, this gain does not translate into more reliable or faithful reasoning trajectories. Its \(\mathrm{RC\text{-}Acc}\) peaks early at around \(51.20\%\) between 20 and 30 training steps, but then declines to \(47.36\%\) by step 140. This deterioration is driven by a substantial increase in reasoning-answer inconsistency during the later stages of training: \(\mathrm{IncR}\) rises from \(6\%\) at step 60 to \(19\%\) at step 140, while \(\mathrm{CBIR}\) increases steadily from \(2\%\) to \(6\%\) over the full training process. These results suggest that while outcome-only RL can improve final-answer correctness, it simultaneously degrades reasoning reliability in multimodal reasoning tasks.

\textbf{PRM offers only limited benefits for reducing reasoning-answer inconsistency.}
Although PRM has been reported to be effective for some text-only reasoning tasks, incorporating a PRM reward does not yield a substantial improvement in \(\mathrm{RC\text{-}Acc}\) under our multimodal reasoning setup. Instead, its trend resembles that of vanilla RLVR: \(\mathrm{RC\text{-}Acc}\) plateaus during the middle stage of training and then declines afterward. PRM-based training is able to prevent inconsistency rates from growing markedly, but it does not reduce them in a meaningful way either. This finding suggests that PRMs remain constrained by their generalization ability, which is a common criticism of such approaches, because the policy model can generate responses that differ substantially from the PRM's training distribution.

\textbf{Our group-wise ranking reward is the most effective at reducing inconsistent reasoning trajectories.}
It is the only method that delivers stable and continuous improvement across all four panels: it consistently improves both answer accuracy and \(\mathrm{RC\text{-}Acc}\), while simultaneously reducing \(\mathrm{IncR}\) and \(\mathrm{CBIR}\). These results indicate that our method provides more stable and reliable reward signals. More importantly, they offer clear evidence that our generative reward alters the direction of optimization, rather than merely stacking training compute.

%% file: parts/4_exp/4_3b_model_size.tex
\begin{figure}[t]
\centering
\includegraphics[width=\columnwidth]{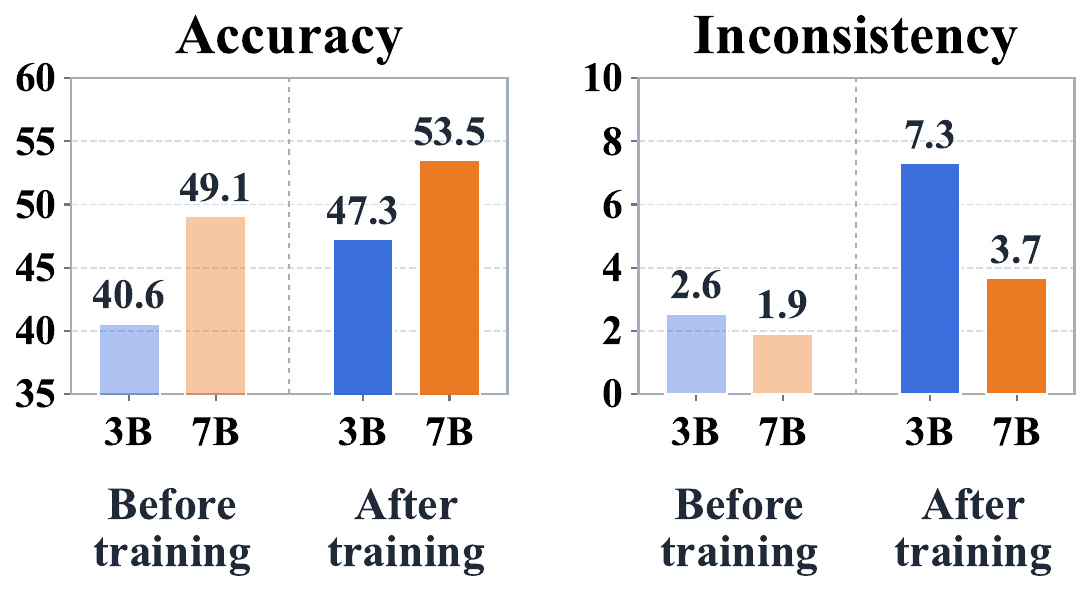}
\caption{Qwen2.5-VL-3B and 7B models trained with vanilla RLVR: Accuracy (left) and Inconsistency (right). Here, Inconsistency denotes the correct-but-inconsistent rate ($\mathrm{CBIR}$). In each panel, the first two bars show the initial checkpoint and the last two bars show the final checkpoint. All values are percentages.}
\label{fig:model_capacity_inconsistency}
\end{figure}

\subsubsection{Effect of Model Size on Inconsistency}
We next ask whether the significance of the reasoning–answer inconsistency problem is related to the size of the model. Figure~\ref{fig:model_capacity_inconsistency} compares Qwen2.5-VL 3B and 7B checkpoints under the same ViRL training setup. We find that \textbf{scaling mitigates the model's inconsistency, but it does not remove the incentive that creates it.} The 7B model starts from a lower correct-but-inconsistent rate than the 3B model, but both models still drift upward under outcome-only RL. The result suggests that model capacity changes the severity of the failure mode, whereas the underlying cause remains the reward signal.

%% file: parts/4_exp/4_3c_mapping_function.tex
\subsubsection{Effect of the Rank-to-Score Mapping Function}
\begin{figure}[t]
\centering
\includegraphics[width=\columnwidth]{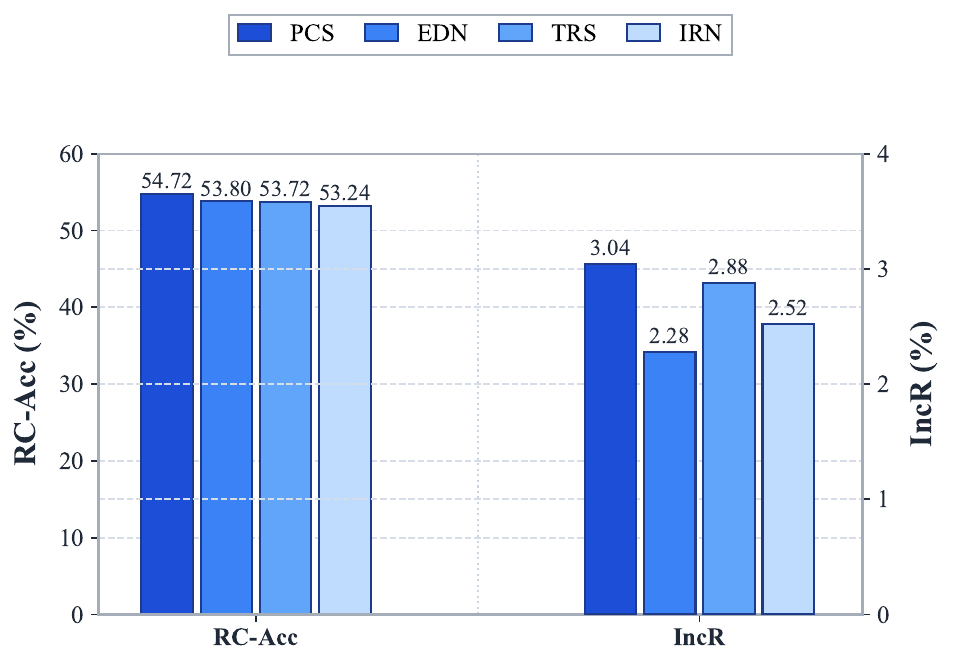}
\caption{Ablation on the rank-to-score mapping function under the matched training setup. The left group reports \(\mathrm{RC\text{-}Acc}\) with the left y-axis, and the right group reports \(\mathrm{IncR}\) with the right y-axis. Bars are ordered as PCS (default), EDN, TRS, and IRN.}
\vspace{-0.7cm}
\label{fig:mapping_ablation}
\end{figure}

After the judge produces an ordering over verifier-passed responses, we still need a scalar mapping from that ordering to rollout reward. We compare four simple options. \textbf{Pairwise-Comparison Score (PCS)}, our default mapping, converts the same \(K\)-candidate ranking into a tie-aware pairwise win rate: each candidate gets \(1\) point against every lower-ranked verifier-passed candidate, \(0.5\) against every tied candidate, and \(0\) against every higher-ranked candidate, then normalizes by \(K-1\). 
The three alternatives are \textbf{Exponential-Decay Normalization (EDN)}, which uses an exponentially decaying rank score before normalization; \textbf{Tiered-Rank Score (TRS)}, which maps each final rank tier directly to evenly spaced linear rewards; and \textbf{Inverse-Rank Normalization (IRN)}, which assigns each candidate a reciprocal score \(1/r\) and then min-max normalizes it within the prompt. Figure~\ref{fig:mapping_ablation} compares these four mappings using the two most relevant summary metrics for this design choice: five-benchmark average \(\mathrm{RC\text{-}Acc}\) and \(\mathrm{IncR}\).

As the figure suggests, \textbf{the main gain comes from group-wise comparison itself, not from a fragile post-ranking mapping.} Across the four mappings, average \(\mathrm{RC\text{-}Acc}\) varies only from \(53.24\%\) to \(54.72\%\), a spread of \(1.48\) points. PCS performs the best, but the performance differences among the four variants are still minor. This shows that once the judge can reliably compare correct candidates within a prompt, the exact monotonic rank-to-rewards mapping matters much less.

%% file: parts/4_exp/4_3d_stability.tex
\subsubsection{Stability Comparison Between Pointwise GRs and Groupwise Ranking}
Groupwise Ranking Reward is motivated by the fact that all rollouts for the same prompt are judged together in a single pass under a shared comparison standard. In contrast, pointwise GR scores each response independently, so judging \(N\) rollouts requires \(N\) separate judge calls, and the judge may apply slightly different internal standards across those calls. Therefore, we assume that Groupwise Ranking Reward has a stability advantage over pointwise GR.
We evaluate this stability through a controlled simulation: if a judge is stable, repeated calls on the same fixed responses should produce similar rewards.

Specifically, we randomly sample \(500\) prompts, generate \(8\) rollouts for each prompt as in the training setting, and rerun the reward assignment process \(4\) times at a relatively low temperature (\(0.7\)). In each repeat, we construct rewards exactly as in training: pointwise GR scores each response independently, whereas Groupwise Ranking Reward jointly ranks the verifier-passed responses within the same prompt and maps that ranking to scalar rewards. We then compute the variance of the repeated reward values and average this quantity over the full sampled set. This metric measures how much the assigned reward changes under small perturbations; lower values therefore indicate more stable reward assignment. We find that \textbf{groupwise ranking yields substantially lower reward variance.} The mean variance drops from \(\mathbf{0.026}\) under pointwise GR to \(\mathbf{0.016}\) under Groupwise Ranking Reward, a \(\mathbf{37.6\%}\) reduction. This suggests that relative comparison makes the assigned reward less sensitive to judge randomness and provides a more stable supervision signal.

%% file: parts/5_conclude.tex
\section{Conclusion}

We study reasoning-answer inconsistency in multimodal RLVR and show that outcome-only optimization can increase this mismatch even when answer accuracy improves. We compare trajectory supervision methods and find that they mitigate this effect. More broadly, these results suggest that answer correctness alone is an incomplete objective for multimodal reasoning, and that trajectory-level supervision remains valuable even after verifiable rewards already improve end accuracy. Among them, \textbf{Groupwise Ranking Reward} performs best by jointly ranking verifier-passed trajectories for the same prompt. It achieves the strongest balance between accuracy and faithfulness, improving average accuracy from 53.6\% to 55.9\% and reliability-conditioned accuracy from 47.4\% to 54.7\% over RLVR, while requiring less judge computation than pointwise generative rewards. 
We also find that groupwise ranking is more stable than pointwise GR, suggesting that shared within-prompt comparison provides a cleaner optimization signal.

\section*{Limitations}

Our study focuses on a specific phenomenon that becomes visible in answer-verifiable RLVR settings: the model can receive reward for a correct final answer even when the reasoning trajectory does not actually support it. This framing fits domains where correctness can be checked reliably, but it does not directly cover less verifiable generation settings without a single canonical answer, such as creative writing, free-form visual dialogue, subjective caption revision, or brainstorming tasks. In such settings, the same failure mode may not appear in the same explicit form, and both inconsistency measurement and reward design may need to be reformulated. Extending the analysis to these less verifiable multimodal tasks is an important direction for future work.

%% file: parts/6_appendix.tex
\section{Standard GRPO Objective}
\label{app:standard_grpo}

For completeness, we summarize the standard GRPO objective used in DeepSeekMath and later RLVR systems \citep{shao2024deepseekmath,guo2025deepseekr1}. Given rollout rewards $\{\mathcal{R}_i\}_{i=1}^N$ for one sampled group, GRPO first standardizes them within the prompt:
\begin{equation}
    A_i = \frac{\mathcal{R}_i - \bar{\mathcal{R}}}{\operatorname{std}(\{\mathcal{R}_j\}_{j=1}^N) + \epsilon_A},
    \qquad
    \bar{\mathcal{R}} = \frac{1}{N}\sum_{j=1}^N \mathcal{R}_j,
\end{equation}
where $\epsilon_A$ is a small constant for numerical stability. Let
\begin{equation}
    \rho_i(\theta) = \frac{\pi_\theta(y_i \mid x)}{\pi_{\theta_{\text{old}}}(y_i \mid x)}.
\end{equation}
The clipped GRPO surrogate is
\begin{equation}
    \resizebox{\columnwidth}{!}{$
    \begin{aligned}
        \mathcal{L}_{\text{GRPO}}(\theta)
        &=
        \mathbb{E}_{x \sim \mathcal{D}, \{y_i\}_{i=1}^N \sim \pi_{\theta_{\text{old}}}(\cdot \mid x)}
        \Bigg[
        \frac{1}{N}\sum_{i=1}^N
        \ell_{\text{clip}}(\theta; x, y_i, A_i) \\
        &\qquad
        - \beta \mathbb{D}_{\mathrm{KL}}\!\left(\pi_\theta(\cdot \mid x)\,\|\,\pi_{\mathrm{ref}}(\cdot \mid x)\right)
        \Bigg]
    \end{aligned}
    $}
\end{equation}
where
\begin{equation}
    \resizebox{\columnwidth}{!}{$
    \ell_{\text{clip}}(\theta; x, y_i, A_i)
    =
    \min\Bigl(
        \begin{aligned}
            &\rho_i(\theta) A_i, \\
            &\operatorname{clip}(\rho_i(\theta), 1-\epsilon, 1+\epsilon) A_i
        \end{aligned}
    \Bigr)
    $}
\end{equation}
Our method does not modify this optimizer; it only changes how $\mathcal{R}_i$ is constructed.

\section{More Details of Groupwise Ranking Reward}
\label{Groupwise_Ranking_Reward}

\paragraph{Groupwise Ranking Reward.}
For the same prompt, let $I^+=\{i:r_{\text{ver}}(y_i)=1\}$ be the verifier-passed set and let $K=|I^+|$. The groupwise judge reads all textualized candidates $\{t_i\}_{i\in I^+}$ together and returns ordered tie-aware tiers $(\tau_1,\dots,\tau_M)$. We then apply exactly the tie-aware pairwise-win-rate mapping and within-group centering defined in Section~\ref{subsec:judge}; if $K<2$, the auxiliary reward is set to zero. This is only a post-processing of the same \(K\)-way ranking, not a separate round of pairwise judge calls. A candidate is treated as beating every lower tier, tying candidates in its own tier, and losing to every higher tier. For example, if \(K=4\) and the judge outputs \(\tau_1=\{A\}\), \(\tau_2=\{B,C\}\), and \(\tau_3=\{D\}\), then the raw scores before centering are \(\tilde{s}_A=1\), \(\tilde{s}_B=\tilde{s}_C=0.5\), and \(\tilde{s}_D=0\). If all verifier-passed rollouts tie, the centered reward becomes zero for all of them.


\subsection{Reward Judge Instructions}
\label{app:reward_judge_instructions}

To make the controlled comparison concrete, we summarize the prompt families used by the three trajectory-supervision variants. The PRM follows a visual stepwise process-reward setup, while the two GR-based variants use the same \texttt{gpt-oss-20b} judge backbone with different user instructions.

\paragraph{PRM instruction.}
For each trajectory step, the PRM uses the following prompt pair and asks the model to emit exactly one token for the current step:
{\nolinenumbers
\begin{promptbox}{PRM Prompt}
\promptrole{System}
You are a process reward model. Given the current reasoning step and prior
context, output exactly one token: \texttt{+} or \texttt{-}. Use \texttt{+}
only if the current step is valid and consistent with the question/context.

\promptdivider

\promptrole{User}

\promptsection{Question}
\begin{promptinputblock}
\{question\}
\end{promptinputblock}

\promptsection{Solution Process}
\begin{promptinputblock}
\{step\}
\end{promptinputblock}
\end{promptbox}
}
When the trajectory has multiple steps, previous steps are replayed as earlier user turns paired with positive assistant replies, and the current step is queried last. The rollout-level PRM reward is then obtained by averaging the per-step rewards.

\paragraph{Pointwise GR instruction.}
Pointwise GR uses an instruction that asks the judge to score one textualized trajectory independently:
{\nolinenumbers
\begin{promptbox}{Pointwise GR Instruction}
You are an expert evaluator for reasoning quality.
Evaluate the assistant response based on reasoning-quality assessment.

\promptdivider

\promptsection{Scoring Rubric}
Score the overall reasoning quality on a scalar in \texttt{[0, 1]}, where:
\begin{itemize}[leftmargin=1.25em, itemsep=2pt, topsep=2pt]
    \item \texttt{1.0}: clear, logically coherent, and mathematically sound derivation.
    \item \texttt{0.8}: mostly solid reasoning with only minor slips that do not break the core logic.
    \item \texttt{0.5}: partially valid reasoning; noticeable gaps or mistakes reduce reliability.
    \item \texttt{0.2}: largely flawed reasoning with major logical/calculation issues.
    \item \texttt{0.0}: reasoning is missing, nonsensical, or clearly fails to solve the task.
\end{itemize}

\promptdivider

\promptsection{Evaluation Guidance}
When assigning the single score, consider the overall logical coherence of
the steps, correct use of given conditions or visual facts, calculation and
transformation quality, clarity/completeness of the conclusion, and whether
uncertainty is handled honestly without unsupported guessing.

\textbf{Note.} The original problem may include image input, but you cannot
access the image. If the reasoning references information that could
plausibly come from the image and is not contradicted by available text, do
not treat that as an error by itself; score based on textual reasoning
quality.

\promptdivider

\promptsection{Output Schema}
Return JSON only with this schema:
\begin{promptjsonbox}
\{"reasoning\_feedback":\par
\ \ "<short explanation of strengths, weaknesses, and key issues>",\par
\ "judge\_score": <float between 0 and 1>\par
\}
\end{promptjsonbox}
Generate \texttt{"reasoning\_feedback"} first and place
\texttt{"judge\_score"} last.

If some information is missing, still provide the best-effort score from
available text.

\promptdivider

\promptsection{Problem}
\begin{promptinputblock}
\{problem\}
\end{promptinputblock}

\promptsection{Response}
\begin{promptinputblock}
\{response\}
\end{promptinputblock}
\end{promptbox}
}

\paragraph{Groupwise Ranking Reward instruction.}
The groupwise variant uses the same judge backbone but replaces scalar scoring with a tie-aware ranking instruction over all verifier-passed candidates for the same prompt:
{\nolinenumbers
\begin{promptbox}{Groupwise Ranking Instruction}
You are ranking multiple candidate solutions to the same geometry problem.

\promptsection{Problem}
\begin{promptinputblock}
\{problem\}
\end{promptinputblock}

\promptsection{Reference Answer}
\begin{promptinputblock}
\{reference\_answer\}
\end{promptinputblock}

\promptsection{Candidate Solutions}
\begin{promptinputblock}
\{candidate\_solutions\}
\end{promptinputblock}

\promptdivider

\promptsection{Ranking Principles}
\begin{enumerate}[leftmargin=1.4em, itemsep=2pt, topsep=2pt]
    \item Audit the reasoning steps, not just the final answer; flag leaps, contradictions, or missing justification.
    \item Prioritise solutions whose final answers align with the reference answer.
    \item If solutions are equally correct and use materially similar reasoning, give them the same rank; do not force an ordering without a clear qualitative difference. If all are equally correct, all ranks must be 1.
    \item When reasoning differs, choose the mathematically valid, well-supported argument even if multiple solutions reach the same final answer.
    \item Penalise calculation mistakes, invalid assumptions, or logical gaps relative to error-free solutions.
    \item Keep the comparison focused on mathematical correctness and clarity.
\end{enumerate}

\promptdivider

\promptsection{Output Format}
Return \textbf{strict JSON} with no commentary:
\begin{promptjsonbox}
\{"solutions": [\par
\ \ \{\par
\ \ \ "index": 1,\par
\ \ \ "rank": 1,\par
\ \ \ "justification": "Short comparison that explains the placement.",\par
\ \ \ "agreement\_with\_reference":\par
\ \ \ \ "match" | "different" | "unknown",\par
\ \ \ "errors": ["optional list of key mistakes"]\},\par
\ \ ...\par
\ ]\}
\end{promptjsonbox}

\promptsection{Rules}
\begin{itemize}[leftmargin=1.25em, itemsep=2pt, topsep=2pt]
    \item Lower rank numbers correspond to better solutions.
    \item Use the same rank for tied solutions that should be treated equally.
    \item Include every candidate exactly once.
    \item If you assign the same rank, the numeric \texttt{rank} fields must be identical (e.g., all 1s for a full tie); do \emph{not} output \texttt{1,2,3...} when you describe them as equal.
    \item Keep the justification concise, citing key reasoning differences or errors that justify the rank.
\end{itemize}
\end{promptbox}
}

\section{Inconsistency Judge Design and Reliability}
\label{app:judge_reliability}


\paragraph{Qualitative manual inspection and observed failure modes.}
We further conducted targeted manual inspection of the inconsistency judgments across benchmarks to understand the judge's remaining failure cases. These inspections suggest that the problematic cases concentrate in a small number of recurring patterns rather than arbitrary misclassifications. The first is \emph{image-grounded option semantics}. Some multiple-choice problems place the semantic content of the options in the image itself, while the textual prompt only exposes letters such as A/B/C/D. In those cases, a text-only judge may see that the reasoning discusses a concrete visual choice while the boxed answer contains only a letter, and it can incorrectly flag the sample as inconsistent because the letter-to-content mapping is visually hidden. The second is \emph{implicit answer revision}. Some responses mention an early provisional answer, then later switch to a different boxed answer with only a terse correction or an incomplete explanation. Humans often read the later resolution as the model's true conclusion, but a strict extractor may attach to the earlier explicit statement and mark the sample inconsistent. These observations clarify the main residual gap between automatic judging and human inspection, and suggest that future improvements should focus on better handling visually grounded option mappings and late-stage answer revisions.

\section{Detailed Controlled Comparison}
\label{app:detailed_controlled_comparison}

Table~\ref{tab:main_step140_detailed} provides the full version of the main controlled comparison. Each cell reports the three-number breakdown used throughout the paper: \(\mathrm{RC\text{-}Acc}\), \(\mathrm{Acc}\), and \(\mathrm{CBIR}\). We place this detailed table in the appendix so the main text can emphasize the faithfulness-aware metric directly, while still making the complete accuracy-versus-inconsistency trade-off explicit.

\begin{table*}[t]
\centering
\caption{Detailed controlled comparison at global step 140. Each cell reports \scalebox{0.84}{[\,\shortstack[c]{{Strict Accuracy}\\{\footnotesize Standard Accuracy / Correct-but-Inconsistent Rate}}\,]}. We bold the best Strict Accuracy in each column and underline the second best.}
\small
\setlength{\tabcolsep}{4.4pt}
\renewcommand{\arraystretch}{1.15}
\begin{tabular}{lcccccc}
\toprule
\textbf{Method}
& \textbf{Avg.}
& \shortstack[c]{\textbf{Math}\\\textbf{Vision}}
& \shortstack[c]{\textbf{Math}\\\textbf{Vista}}
& \textbf{MMMU}
& \shortstack[c]{\textbf{MMMU-}\\\textbf{Pro}}
& \textbf{WeMath} \\
\midrule
\multicolumn{7}{c}{\emph{Reference checkpoints}} \\
\midrule
\methodcell{Qwen2.5-VL-7B-IT}
& \resultcell{47.3}{49.0}{1.7}
& \resultcell{24.8}{26.2}{1.4}
& \resultcell{67.2}{68.6}{1.4}
& \resultcell{52.6}{55.0}{2.4}
& \resultcell{35.2}{37.4}{2.2}
& \resultcell{56.6}{58.0}{1.4} \\

\methodcell{RLVR}
& \resultcell{47.4}{53.6}{6.2}
& \resultcell{22.8}{28.6}{5.8}
& \resultcell{70.8}{75.6}{4.8}
& \resultcell{48.2}{55.8}{7.6}
& \resultcell{36.2}{40.4}{4.2}
& \resultcell{58.8}{67.4}{8.6} \\

\midrule
\multicolumn{7}{c}{\emph{Controlled comparison with different reward methods}} \\
\midrule
\methodcell{\textit{w/} PRM}
& \resultcell{46.8}{49.1}{2.3}
& \resultcell{26.4}{26.8}{0.4}
& \resultcell{66.4}{70.0}{3.6}
& \resultcell{50.0}{51.6}{1.6}
& \resultcell{30.6}{35.2}{4.6}
& \resultcell{60.8}{61.8}{1.0} \\

\methodcell{\textit{w/} Pointwise GR}
& \resultcell{\underline{52.3}}{53.8}{1.5}
& \resultcell{\underline{31.0}}{32.4}{1.4}
& \resultcell{\underline{71.0}}{72.2}{1.2}
& \resultcell{\underline{56.6}}{58.0}{1.4}
& \resultcell{\textbf{39.0}}{40.0}{1.0}
& \resultcell{\underline{63.8}}{66.2}{2.4} \\

\methodcell{\textit{w/} Groupwise Ranking Reward}
& \resultcell{\textbf{54.7}}{55.9}{1.2}
& \resultcell{\textbf{31.4}}{33.2}{1.8}
& \resultcell{\textbf{74.8}}{76.4}{1.6}
& \resultcell{\textbf{57.8}}{58.6}{0.8}
& \resultcell{\underline{38.8}}{39.6}{0.8}
& \resultcell{\textbf{70.8}}{71.6}{0.8} \\
\bottomrule
\end{tabular}
\label{tab:main_step140_detailed}
\end{table*}

\section{More Results on Factors Affecting Inconsistency}
\label{app:inconsistency_factors}

\subsection{Question Type and Training Objective}
\label{app:question_type_breakdown}

The five-benchmark evaluation suite is not balanced by question type, so raw inconsistency counts are hard to interpret. Across the 2{,}500 evaluated problems, 1{,}990 are multiple-choice and 510 are open-ended. The open-ended subset appears only in MathVision (247), MathVista (233), and a small 30-example subset of MMMU, while MMMU-Pro and WeMath are entirely multiple-choice. We therefore normalize every rate in this subsection by the total number of examples within the same question type rather than by the global benchmark total.

\begin{table*}[t]
\centering
\caption{Question-type breakdown under three training settings. Every percentage is normalized by the total number of examples within the same question type, not by the global benchmark total.}
\small
\setlength{\tabcolsep}{5.2pt}
\renewcommand{\arraystretch}{1.12}
\resizebox{\textwidth}{!}{%
\begin{tabular}{lcccccccc}
\toprule
\textbf{Training Setup}
& \multicolumn{4}{c}{\textbf{Multiple-choice} (\(n=1{,}990\))}
& \multicolumn{4}{c}{\textbf{Open-ended} (\(n=510\))} \\
\cmidrule(lr){2-5} \cmidrule(lr){6-9}
& \textbf{StdAcc} & \textbf{StrictAcc} & \textbf{IncR} & \textbf{CBIR}
& \textbf{StdAcc} & \textbf{StrictAcc} & \textbf{IncR} & \textbf{CBIR} \\
\midrule
Before training & 51.76\% & 49.40\% & 5.88\% & 2.36\% & 38.82\% & 38.63\% & 1.76\% & 0.20\% \\
RLVR & 55.83\% & 48.84\% & 20.40\% & 6.98\% & 44.71\% & 41.57\% & 14.71\% & 3.14\% \\
\textit{w/} Groupwise Ranking Reward & 58.74\% & 57.39\% & 3.12\% & 1.36\% & 44.71\% & 44.31\% & 2.75\% & 0.39\% \\
\bottomrule
\end{tabular}
}
\label{tab:question_type_inconsistency}
\end{table*}

\begin{figure}[t]
\centering
\includegraphics[width=\columnwidth]{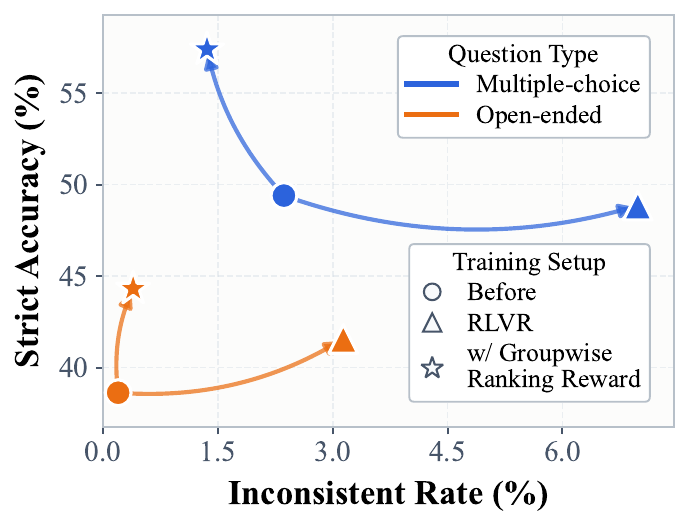}
\caption{Question-type trade-off trajectories under three training settings. Color denotes question type and marker denotes training setup. For each question type, both trained checkpoints are connected back to the shared before-training model, so the figure shows how RLVR and \textit{w/} Groupwise Ranking Reward move the model in strict-accuracy versus inconsistency space. The panel plots \(\mathrm{StrictAcc}\) against Inconsistent Rate, where Inconsistent Rate denotes the correct-but-inconsistent rate. Every percentage is normalized within the same question type.}
\label{fig:question_type_inconsistency}
\end{figure}

\textbf{Training objective matters more than the raw question-type split.} Table~\ref{tab:question_type_inconsistency} and Figure~\ref{fig:question_type_inconsistency} show that the low open-ended inconsistency of the reference checkpoint does not survive outcome-only RL. Before training, \(\mathrm{IncR}\) is \(5.88\%\) for multiple-choice and \(1.76\%\) for open-ended. Under RLVR, these rates jump to \(20.40\%\) and \(14.71\%\), and \(\mathrm{CBIR}\) rises to \(6.98\%\) and \(3.14\%\). Question type still changes the absolute level, but the dominant effect is the reward design: outcome-only training pushes both question types toward much larger reasoning-answer mismatch.

\textbf{Groupwise Ranking Reward recovers faithfulness without giving up answer gains.} On multiple-choice problems, Groupwise Ranking Reward improves \(\mathrm{StdAcc}\) from \(51.76\%\) to \(58.74\%\) while driving \(\mathrm{IncR}\) below the starting checkpoint, from \(5.88\%\) to \(3.12\%\). The open-ended comparison is even cleaner: RLVR and Groupwise Ranking Reward reach the same \(\mathrm{StdAcc}\) of \(44.71\%\), yet Groupwise Ranking Reward cuts \(\mathrm{IncR}\) from \(14.71\%\) to \(2.75\%\) and \(\mathrm{CBIR}\) from \(3.14\%\) to \(0.39\%\), which lifts \(\mathrm{StrictAcc}\) from \(41.57\%\) to \(44.31\%\). In other words, the open-ended improvement comes almost entirely from better faithfulness rather than better Standard Accuracy.

\textbf{The qualitative failure mode still differs by question type.} For multiple-choice problems, the dominant pattern is still option-selection drift: the reasoning converges to one content answer, but the final boxed output switches to a different option letter or to a nearest-choice guess. For open-ended problems, the mismatch is more often a late revision or answer-type collapse: the model derives one free-form result and then boxes a different scalar, count, or endpoint without a new derivation. These mechanisms differ, but both are strongly amplified by RLVR and sharply reduced by Groupwise Ranking Reward.

\subsection{Training Data Distribution: ViRL vs.\ GSM8K}
\label{app:cross_dataset_training}

A second factor is the RL training data distribution. To study it, we compare two runs that start from the same initialization and are evaluated with the same five-benchmark multimodal inconsistency pipeline, but are trained on two very different data sources: a multimodal image-grounded RL corpus \textit{ViRL} \citep{wang2025vlrethinker}, and \textit{GSM8K}, a pure text arithmetic dataset \citep{cobbe2021gsm8k}. Both runs use the same GRPO training framework, but the underlying task distributions and correctness signals are not identical across the two datasets. We therefore present this as an empirical cross-dataset comparison rather than a clean modality-only causal ablation.

\begin{figure}[t]
\centering
\includegraphics[width=\columnwidth]{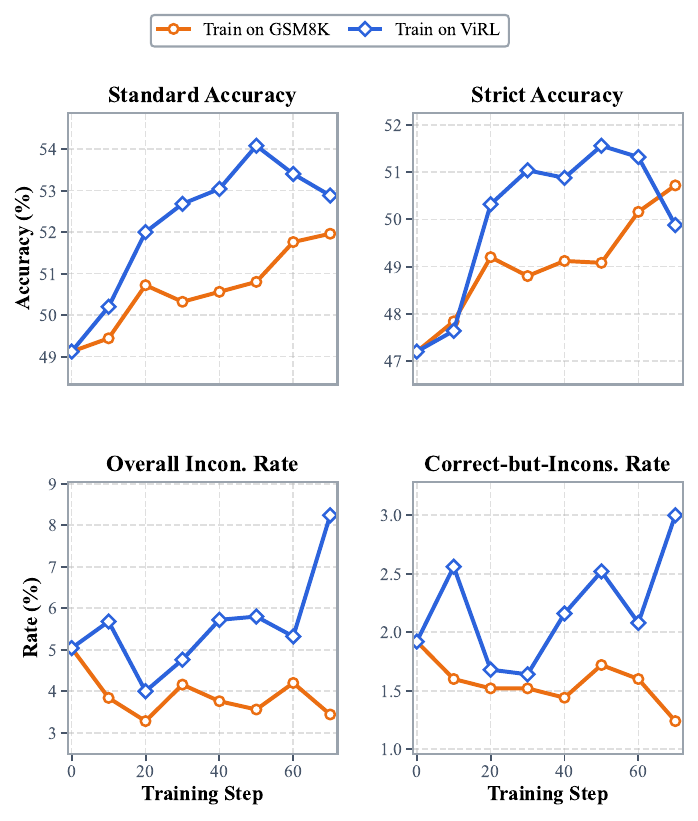}
\caption{Average evaluation dynamics when the same RL framework is trained on text-only GSM8K or multimodal ViRL. Both runs start from the same initialization and are evaluated with the same five-benchmark multimodal inconsistency pipeline. ViRL reaches higher answer accuracy, but its inconsistency grows much faster during training.}
\label{fig:cross_dataset_training_curves}
\end{figure}

\textbf{Figure~\ref{fig:cross_dataset_training_curves} shows that GSM8K keeps inconsistency low while still improving multimodal evaluation.} Relative to the shared initialization, training on GSM8K improves average Standard Accuracy from \(49.12\%\) to \(51.96\%\) and Strict Accuracy from \(47.20\%\) to \(50.72\%\). At the same time, overall inconsistency drops from \(5.04\%\) to \(3.44\%\), and the correct-but-inconsistent rate drops from \(1.92\%\) to \(1.24\%\). Across training, the inconsistency curves remain in a relatively tight band and stay well below the shared initialization. This means that text-only RL data can still transfer useful reasoning discipline to multimodal evaluation, even though it does not directly train image-grounded skills.

\textbf{ViRL offers higher upside, but the faithfulness trade-off is much less stable.} On the same evaluation pipeline, ViRL reaches higher answer accuracy and a stronger best \(\mathrm{RC\text{-}Acc}\) than GSM8K. However, this gain is accompanied by substantially higher inconsistency, and the later part of training degrades further rather than remaining stable. At the end of training, ViRL still keeps Standard Accuracy above the shared initialization at \(52.88\%\), but overall inconsistency rises to \(8.24\%\) and \(\mathrm{RC\text{-}Acc}\) falls to \(49.88\%\). The same pattern appears broadly across benchmarks rather than in one isolated dataset: at the final checkpoints, ViRL has higher inconsistency than GSM8K on all five evaluation sets, for example \(10.0\%\) vs.\ \(3.0\%\) on MathVision and \(7.8\%\) vs.\ \(3.2\%\) on WeMath.

\textbf{The training logs support the same asymmetry.} We also inspect the judge-flagged inconsistent samples saved from training-time logs. On GSM8K, these cases are nearly absent: only \(0.0\%-0.4\%\) of the 500 inspected rollouts per checkpoint are flagged inconsistent, and high-score inconsistent samples never exceed \(0.2\%\). On ViRL, the corresponding numbers are much larger, ranging from \(6.2\%\) to \(10.8\%\) for all flagged inconsistencies and from \(2.4\%\) to \(5.6\%\) for inconsistent samples with score \(>0.5\). This does not prove that multimodality alone causes inconsistency, but it does show that the ViRL training distribution exposes the policy to many more internally mismatched yet nontrivially rewarded rollouts, which is consistent with the evaluation-side drift in Figure~\ref{fig:cross_dataset_training_curves}.

Overall, this comparison suggests that the training data distribution has a first-order effect on faithfulness dynamics under RL. GSM8K behaves like a conservative regularizer: it transfers some answer-format and reasoning discipline, keeps inconsistency low, and improves Strict Accuracy steadily, but its multimodal accuracy ceiling is limited. ViRL has a higher ceiling on multimodal tasks, yet it also produces a much stronger late-stage inconsistency drift. The practical implication is not that text-only data is universally preferable, but that multimodal RL appears to need stronger inconsistency control if we want to keep its accuracy gains from turning into reasoning-answer mismatch.